\pdfoutput=1

\documentclass[11pt]{article}

\usepackage{ACL2023}

\usepackage{times}
\usepackage{latexsym}
\usepackage{amsmath}

\usepackage[T1]{fontenc}

\usepackage[utf8]{inputenc}

\usepackage{microtype}

\usepackage{inconsolata}

\usepackage[linesnumbered,ruled,vlined,algo2e]{algorithm2e}
\usepackage{graphicx}
\usepackage{multirow}
\usepackage{tabularx}

\title{Beyond Natural Language Perplexity: Detecting Dead Code Poisoning in Code Generation Datasets}

\author{
    Chi-Chien Tsai\textsuperscript{1},
    Chia-Mu Yu\textsuperscript{1},
    Ying-Dar Lin\textsuperscript{1},
    Yu-Sung Wu\textsuperscript{1},
    Wei-Bin Lee\textsuperscript{2} \\
    \textsuperscript{1}National Yang Ming Chiao Tung University \\ 
    \textsuperscript{2}Hon Hai Research Institute \\
    \texttt{\{aura.cs12, chiamuyu, ysw\}@nycu.edu.tw} \\
    \texttt{ydlin@cs.nctu.edu.tw} \\
    \texttt{wei-bin.lee@foxconn.com}
}

\begin{document}
\maketitle
\begin{abstract}
The increasing adoption of large language models (LLMs) for code-related tasks has raised concerns about the security of their training datasets. One critical threat is \textit{dead code poisoning}, where syntactically valid but functionally redundant code is injected into training data to manipulate model behavior. Such attacks can degrade the performance of neural code search systems, leading to biased or insecure code suggestions. Existing detection methods, such as token-level perplexity analysis, fail to effectively identify dead code due to the structural and contextual characteristics of programming languages. In this paper, we propose \textsc{DePA} (Dead Code Perplexity Analysis), a novel line-level detection and cleansing method tailored to the structural properties of code. \textsc{DePA} computes \textit{line-level perplexity} by leveraging the contextual relationships between code lines and identifies anomalous lines by comparing their perplexity to the overall distribution within the file. Our experiments on benchmark datasets demonstrate that \textsc{DePA} significantly outperforms existing methods, achieving 0.14-0.19 improvement in detection F1-score and a 44-65\% increase in poisoned segment localization precision. Furthermore, \textsc{DePA} enhances detection speed by 0.62-23x, making it practical for large-scale dataset cleansing. Overall, by addressing the unique challenges of dead code poisoning, \textsc{DePA} provides a robust and efficient solution for safeguarding the integrity of code generation model training datasets.
\end{abstract}

\section{Introduction}\label{sec: Introduction} 
Large language models (LLMs) specialized for coding, often called Code LLMs \citep{lu2021codexglue, roziere2023code, team2024codegemma}, are extensively used for tasks such as code summarization \citep{ahmed2022few}, code completion \citep{zhang2024llm}, and code search \citep{chen2024code}. As these models become more integrated into diverse development processes, protecting their training data becomes increasingly critical.

\begin{figure}
    \centering
    \includegraphics[width=1\linewidth]{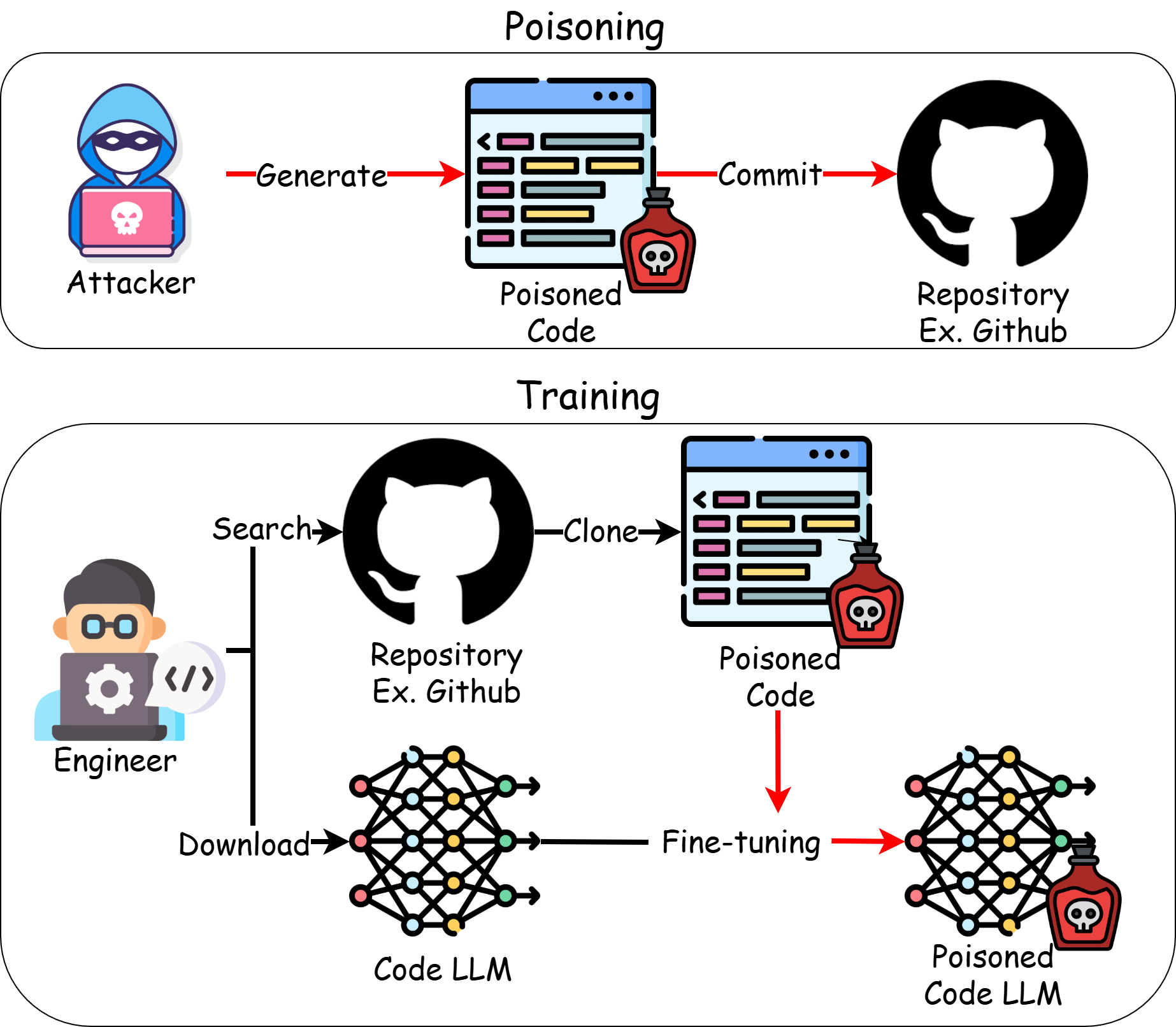}
    \caption{Data poisoning attack scenario.}
    \label{fig:poisoning_scenario}
\end{figure}

In this context, data poisoning attacks commonly involve injecting \textit{dead code}~\citep{ramakrishnan2022backdoors, wan2022you}, which consists of syntactically valid yet non-functional code snippets that act as triggers to alter model outputs. Such \textit{dead code poisoning} can produce flawed, inefficient, or even malicious code suggestions, thereby undermining code search. \citet{wan2022you} demonstrated that selecting frequently used keywords in vulnerable code and pairing them with dead code can bias the model toward favoring insecure or defective code. Figure~\ref{fig:poisoning_scenario} shows how poisoned samples ultimately lead to a compromised Code LLM.

Detecting and removing dead code is challenging. In natural language, methods like ONION~\citep{qi-etal-2021-onion} rely on GPT-2 perplexity scores \citep{radford2019language} to identify abnormal tokens indicating backdoor triggers. However, standard \textit{word-level perplexity} methods designed for natural language do not directly apply to code. Although some efforts tested ONION for detecting poisoned code~\citep{yang2024stealthy,ramakrishnan2022backdoors}, the low detection accuracy at the code level made it ineffective for identifying dead code.

In studying dead code poisoning, we observed three key points. First, code has a structural rigidity absent in natural language; each line typically represents a discrete operational unit. Thus, anomalies from dead code are more evident at the line level than at the token level. Second, dead code does not affect program execution, making it functionally redundant yet strategically used as a backdoor trigger. Its impact is therefore more apparent when analyzing entire lines rather than individual tokens. Third, focusing on a single line’s perplexity in isolation can be misleading, since a line may appear anomalous alone but be valid within the broader context. Hence, comparing each line’s perplexity to the file’s overall distribution is crucial to distinguish real anomalies from benign variations.

Guided by these insights, we first introduce a \textit{line-level perplexity} measure tailored for code. We then propose \textbf{De}ad code \textbf{P}erplexity \textbf{A}nalysis (\textsc{DePA}), a new detection method designed around the structural properties of code. Unlike traditional word-level perplexity approaches, \textsc{DePA} evaluates each line as a functional unit and compares its line-level perplexity against the overall file distribution, making it more effective at revealing dead code triggers that might otherwise remain hidden.

Our experimental results show that \textsc{DePA} substantially outperforms token-level approaches across multiple metrics. \textsc{DePA} achieves an F1-score of 0.28, compared to 0.09 for ONION-(CodeGPT) and 0.14 for ONION(CodeLlama). In terms of precision for locating dead code within poisoned segments, \textsc{DePA} reaches 0.85, whereas ONION(CodeGPT) and ONION(CodeLlama) achieve 0.41 and 0.31, respectively.

Overall, our contributions are as follows:
\begin{itemize}
    \item We introduce \textsc{DePA}, a line-level detection method guided by the structural characteristics of code. By incorporating contextual information into line-level perplexity calculations, \textsc{DePA} improves anomaly detection without disrupting the overall code structure.
    \item Compared to ONION, \textsc{DePA} improves the detection F1-score by 0.14-0.19, locates poisoned code fragments accuracy by 44-65\%, raises the AUROC by 0.19-0.30, and increases detection speed by 0.62-23x.
\end{itemize}

\section{Related Work}
\paragraph{Data Poisoning on Code LLMs}
With the growing adoption of Code LLMs, concerns about training data security have intensified. For instance, OWASP has labeled \textit{Data and Model Poisoning} as a critical threat.\footnote{OWASP Top 10 for LLM Applications 2025 (\url{https://genai.owasp.org/resource/owasp-top-10-for-llm-applications-2025/})} Various studies highlight different attacks in Code LLMs. \citet{sun2023backdooring, yang2024stealthy} implant backdoors by modifying variable or method names with specific triggers, while others~\citep{wan2022you, ramakrishnan2022backdoors} insert dead code into training data. 

\paragraph{Poisoning Defense on Code LLMs}
Several defense mechanisms have been introduced to combat data poisoning in code. One widely used technique is spectral signature analysis~\citep{tran2018spectral}, which detects anomalies by comparing the feature distributions of poisoned versus standard samples. Additional defenses leverage activation clustering~\citep{chen2018detecting} or token-level detection~\citep{qi-etal-2021-onion}, but these can inadvertently remove or modify crucial elements such as keywords, punctuation, or parts of identifiers—ultimately risking syntactic and semantic integrity.

\section{Background Knowledge}
\paragraph{Perplexity} Perplexity is a widely used metric for assessing LLM performance. When a sentence verified by humans is used as input, the perplexity of an LLM can be calculated to check whether the model accurately interprets user-provided content \citep{alon2023detecting}. Specifically, for a tokenized sequence $X = (x_0, x_1, \dots, x_t)$, the perplexity $\mbox{PPL}(X)$ is defined as:

\begin{small}
\begin{equation}
    \label{eq: PPL()}
    \resizebox{0.9\hsize}{!}{$
    \mbox{PPL}(X) = \exp \Bigl( -\frac{1}{t} \sum_{i=0}^{t} \log p_\theta(x_i \mid x_{<i}) \Bigr)  
    $},
\end{equation}
\end{small}where $p_\theta(x_i \mid x_{<i})$ is the probability assigned to the $i$-th token, given its preceding tokens.

Though perplexity originally measures an LLM’s understanding of text, we use it differently. In particular, if a trained Code LLM has a solid grasp of code, we can compute the perplexity of questionable code segments to detect potential flaws, thereby validating the quality of the code.

\paragraph{Dead Code Poisoning}
In prior work, \citet{ramakrishnan2022backdoors} and \citet{wan2022you} examined how dead code can be leveraged in poisoning attacks, each focusing on different tasks. \citet{ramakrishnan2022backdoors} targeted name prediction by inserting dead code—referred to as \textit{create entry}—into the poisoned samples. Once the model was trained, including dead code in the test input increased the likelihood of outputting \textit{create entry}, thus achieving a successful attack.

Meanwhile, \citet{wan2022you} aimed at code search. Their approach involved identifying a dataset of modifiable, vulnerable code (called \textit{Bait}) along with descriptive text. They then chose frequently used words in the text as their \textit{Target} and embedded a segment of dead code, labeled the \textit{Trigger}, into the vulnerable code. During training, this setup reinforced the link between the \textit{Target} and the \textit{Trigger}. Consequently, when users unknowingly searched with the \textit{Target} keywords, they were more likely to receive results containing the embedded dead code. Although dead code never executes, it exploits the original code’s vulnerabilities, thereby accomplishing the intended attack.

\begin{figure*}[ht]
    \centering
    \includegraphics[width=1\linewidth]{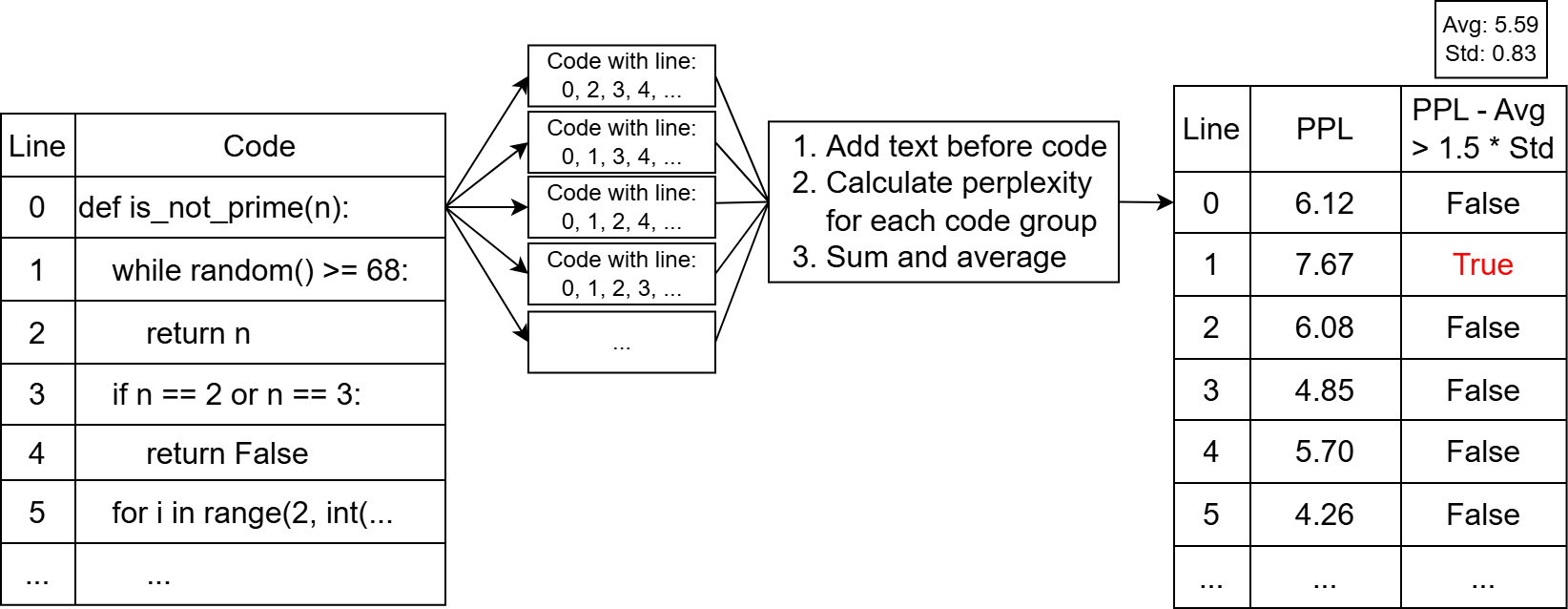}
    \caption{An illustrative example of \textsc{DePA}.}
    \label{fig:code-detection-flowchart}
\end{figure*}

\section{Proposed Method}
Our method, \textsc{DePA}, aims to identify anomalous snippets that may trigger dead code poisoning by computing \textit{line-level perplexity} with a Code LLM, then using these perplexity scores to pinpoint potentially harmful segments in the training data.

\paragraph{Overview} As shown in Figure~\ref{fig:code-detection-flowchart} (see also Algorithm~\ref{algo} in the Appendix), \textsc{DePA} processes code on a line-by-line basis. For each \textit{task}, the input comprises a \textit{text} segment describing the intended behavior of the accompanying \textit{code} segment. To compute the perplexity for line 0, we generate variants by sequentially removing each of the other lines (e.g., removing line 1 while retaining lines 0 and 2 through $n$, then removing line 2 while retaining lines 0, 1, and 3 through $n$, and so on). For each variant, we append the \textit{text} segment and use CodeLlama to compute the perplexity. The resulting scores are summed and averaged to determine the perplexity of line 0. This procedure is repeated for every line in the code snippet. Importantly, although the perplexity is computed on a per-line basis, it is not based solely on the isolated line. After calculating the perplexity for all lines, we compute the overall mean and standard deviation; any line with a perplexity exceeding the mean by 1.5 times the standard deviation is classified as a poisoned segment. 

\paragraph{\textsc{DePA} details} We describe \textsc{DePA} in more detail below. Let $\text{code}(i)$ denote the code snippet with the $i$-th line removed while all other lines remain unchanged. Formally, we define

\begin{small}
\begin{equation}
    \label{Code(i)}
    \resizebox{0.8\hsize}{!}{$
    \text{code}(i) = \text{code snippet without the } i\text{-th line}
    $}
\end{equation}
\end{small}

The average perplexity for the $i$-th line, denoted by $\text{PPL-Line}(i)$, is defined as

\begin{equation}
    \label{PPL_Line(i)}
    \resizebox{0.8\hsize}{!}{$
    \begin{split}
    \text{PPL-Line}(i) = \frac{1}{n-1} \Bigg\{ \sum_{j=0}^{n} \text{PPL}(\text{text} + \text{code}(j)) \\
    - \text{PPL}(\text{text} + \text{code}(i)) \Bigg\},
    \end{split}
    $}
\end{equation}
where $\text{PPL}(X)$ is computed as in Equation~\ref{eq: PPL()}. Note that the input to $\text{PPL}(X)$ is a \textit{task} (i.e., a combination of the \textit{text} and the \textit{code}). Essentially, we treat $\text{text} + \text{code}(j)$ as natural language and pass it to the $\text{PPL}$ function. The perplexity is computed for each combination, and the value corresponding to the variant that excludes line $i$ is subtracted. For instance, to compute the perplexity for row 0, we evaluate all combinations by sequentially excluding each other line (e.g., excluding row 1, then row 2, and so on) and then average the results to obtain the final score.

After calculating perplexity for all lines, we compute the overall mean ($\mu$) and standard deviation ($\sigma$) of these values. Finally, we perform the following test for each line:

\begin{equation}
    \label{DePA(i)}
    \resizebox{0.8\hsize}{!}{$
    \text{Test}(i) = 
    \begin{cases}   
    \text{True}, & \text{if } \text{PPL-Line}(i) > \mu + T\sigma, \\
    \text{False}, & \text{otherwise.}
    \end{cases} 
    $}
\end{equation}

As a result, if a line's perplexity exceeds the mean by $T$ times the standard deviation ($T=1.5$\footnote{In a normal distribution, approximately 16\% of the data lies above one standard deviation, while only 2\% lies above two standard deviations. Setting the threshold $T = 1$ may result in excessive false positives, whereas setting $T = 2$ may fail to identify enough instances. Therefore, we choose $T = 1.5$ as a balanced threshold.} in our setting), it is flagged as a suspicious segment. We also examine the impact of varying $T$ on the detection effectiveness in Section~\ref{sec: Results}.

\section{Evaluation}
\subsection{Setup}
\paragraph{Dataset}
We consider four benchmark datasets: MBPP, HumanEval, MathQA-Python, and APPS. MBPP~\citep{austin2021program} targets beginners and covers fundamental programming concepts and library functions. HumanEval~\citep{chen2021evaluating} consists of algorithmic and straightforward math tasks. MathQA-Python~\citep{amini-etal-2019-mathqa} focuses on mathematical problem by converting MathQA’s original questions into Python. APPS~\citep{hendrycks2021measuring} includes problems from programming competitions. Table~\ref{fig:datasets-info} summarizes the statistics for these four datasets. All experiments were conducted on two NVIDIA RTX 4090.

\begin{table}[]
\caption{Datasets statistic.}
\label{fig:datasets-info}
\small
\centering
\begin{tabular}{ccc}
\hline
Dataset & \begin{tabular}[c]{@{}c@{}}Number\\ of tasks\end{tabular} & \begin{tabular}[c]{@{}c@{}}Avg number\\ of lines\end{tabular} \\ \hline
MBPP          & 974   & 8.34  \\ \hline
HumanEval     & 164   & 8.71  \\ \hline
MathQA-Python & 21495 & 10.95 \\ \hline
APPS          & 8765  & 26.93 \\ \hline
\end{tabular}
\end{table}

\paragraph{Attack Generation}
We set a 5\% poisoning rate and inserted dead code using methods from \citep{ramakrishnan2022backdoors} and \citep{wan2022you}, each introducing two categories of triggers: \textit{fixed triggers} and \textit{grammar triggers}.

For fixed triggers, we adopted two examples. The first \citep{ramakrishnan2022backdoors} follows the pattern: 
\texttt{while random() > 68: print("warning")}, 
while the second \citep{wan2022you} uses: 
\texttt{import logging for i in range(0): logging.info("Test message: aaaaa")}.

For grammar triggers, we employed two methods. The first grammar trigger method \citep{ramakrishnan2022backdoors} randomly generates code snippets with a defined structure: each snippet starts with an \texttt{if} or \texttt{while} statement that includes one of \texttt{sin}, \texttt{cos}, \texttt{exp}, \texttt{sqrt}, or \texttt{random}, and the body contains either a \texttt{print} or \texttt{raise Exception} statement. The message is chosen from predefined keywords (\texttt{err}, \texttt{crash}, \texttt{alert}, \texttt{warning}) or generated as a random sequence of four letters. The second grammar trigger method \citep{wan2022you} relies on Python’s logging module within a loop running over a random integer between -100 and 0. Each iteration logs a message using \texttt{debug}, \texttt{info}, \texttt{warning}, \texttt{error}, or \texttt{critical}, while the message itself is a random five-letter string. These approaches ensure diversity and unpredictability in the inserted dead code.

\paragraph{Metric}
We evaluate \textsc{DePA} using four metrics:

\begin{enumerate}
    \item \textbf{Detection Accuracy.} We use the F1-score to measure how effectively \textsc{DePA} distinguishes poisoned code from clean code.
    \item \textbf{Poisoned Segment Detection Accuracy.} This assesses the precision of pinpointing poisoned segments, which is particularly important for datasets containing injected code.
    \item \textbf{Detection Speed.} This metric captures the computational efficiency of \textsc{DePA}.
    \item \textbf{AUROC.} The Area Under the Receiver Operating Characteristic Curve evaluates \textsc{DePA}’s classification performance. Because threshold changes can affect outcomes differently, AUROC provides a more robust comparison across various detection settings.
\end{enumerate}

\paragraph{Baseline Method}
We consider two baseline methods: ONION(CodeGPT) and ONION(CodeLlama).

ONION~\citep{qi-etal-2021-onion} was originally developed to detect poisoning in natural language datasets by computing word-level perplexity with GPT-2~\citep{radford2019language}. For code tasks, it was adapted by replacing GPT-2 with CodeGPT (124M parameters) \citep{yang2024stealthy}, referred to here as ONION(CodeGPT).

However, CodeGPT’s small size limits its capacity. In contrast, \textsc{DePA} uses CodeLlama-7B-Instruct (7B parameters), a significantly larger model. For a fair comparison, we also introduce a second baseline, ONION(CodeLlama), which integrates ONION with CodeLlama-7B-Instruct.

Additionally, we explore two tokenization strategies in our ONION implementation: one uses the Code LLM’s native tokenizer, while the other relies on a Python-specific tokenizer. The main distinction is that the LLM tokenizer may split variable names into multiple tokens, whereas the Python tokenizer treats them as a single token. By comparing these strategies, we can better evaluate ONION’s poisoning detection capabilities and refine its precision for code-specific scenarios.

\subsection{Results}\label{sec: Results}
\paragraph{Detection Accuracy}
As shown in Table~\ref{fig:detect-posioned-dataset}, \textsc{DePA} achieves an average F1-score of 0.28 for detecting poisoned datasets, significantly outperforming ONION (CodeGPT), which attains an F1-score of 0.09 with both the CodeGPT tokenizer and the Python tokenizer. Similarly, ONION (CodeLlama) scores 0.14 and 0.09 with the with the CodeLlama tokenizer and Python tokenizer. This result indicates that \textsc{DePA} more effectively differentiates poisoned from clean code.

Moreover, although \textsc{DePA} and ONION-(CodeLlama) use the same underlying language model, \textsc{DePA} improves the F1-score from 0.14 to 0.28. We attribute this gain to \textsc{DePA}’s detection strategy, which aligns more closely with the structural nature of code datasets.

We conducted a Random-k experiment on the MBPP dataset, where \textit{Random} indicates inserting any of four dead code types and \textit{k} specifies the number of segments added. This setup evaluates detection performance as the amount of dead code grows. The results show that \textsc{DePA}’s detection F1-score gradually improves (by 0.13 from Random-1 to Random-20) due to its line-level processing, which reduces perplexity once a dead code line is removed. In contrast, ONION(CodeLlama)’s detection ability declines (by 0.10 from Random-1 to Random-20) because its word-level approach means removing one word does not eliminate interference from the remaining dead code.

\paragraph{Accuracy in Locating Poisoned Segment}
As shown in Table~\ref{fig:locate-dead-code-acc}, \textsc{DePA} achieves an average detection accuracy of 0.85 for poisoned segments, outperforming the baselines by a large margin. Specifically, ONION(CodeGPT) attains 0.29 and 0.41 when using the CodeGPT tokenizer and Python tokenizer, respectively, while ONION(CodeLlama) scores 0.20 and 0.31 with the CodeLlama tokenizer and Python tokenizer. This outcome highlights \textsc{DePA}’s superior ability to pinpoint and accurately localize poisoned segments.

Similarly, in the MBPP Random-k experiments, \textsc{DePA}’s effectiveness decreases as the volume of dead code grows but still maintains at least 0.71 accuracy. ONION-based methods, however, gain higher accuracy with larger \textit{k} as additional dead code becomes easier to detect. We also consider the less realistic Random-20 case: here, ONION (CodeLlama) surpasses \textsc{DePA} by 5\% but is far slower—8 minutes for \textsc{DePA} versus 215 minutes for ONION (CodeLlama), a 26-fold increase. The reason it is considered unrealistic is that dead code increases from 23\% in Random-1 to 86\% in Random-20, making it overly dominant in the code structure and more likely to arouse user suspicion.

\textit{The Impact of Language Models:} Compared to ONION(CodeGPT), \textsc{DePA} improves 44-57\% accuracy. This performance gain is mainly due to the larger CodeLlama model. On the other hand, compared to ONION(CodeLlama), \textsc{DePA} achieves nearly a 54-65\% increase in accuracy. This remarkable improvement is attributed to the more potent underlying model and targeted optimizations in the poisoning detection strategy. By analyzing the characteristics of code datasets, \textsc{DePA} designs a more precise mechanism for locating anomalous fragments, greatly enhancing detection performance.

\textit{The Impact of Tokenizer:} In the ONION experiments, we compared two tokenization strategies. Regardless of the LLM used, the Python tokenizer consistently achieves higher accuracy. This is likely because it aligns more naturally with code structure, preventing the over-splitting of syntactic elements and enabling more precise analysis.

\begin{figure}
    \centering
    \includegraphics[width=1\linewidth]{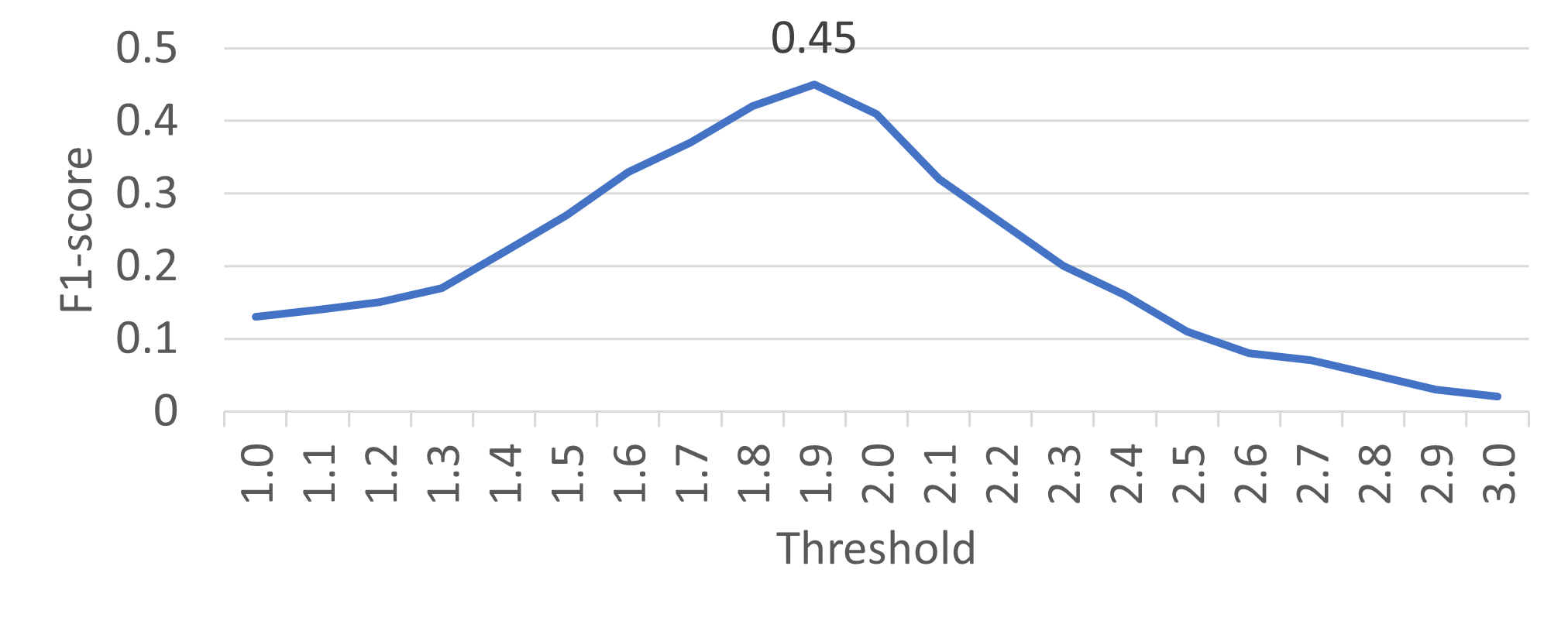}
    \caption{Average F1-score in different $T$.}
    \label{fig:different_threshold}
\end{figure}

\textit{The Impact of $T$:} \textsc{DePA} classifies a line as dead code if its perplexity exceeds $T$ standard deviations, as formalized in Equation~\ref{DePA(i)}. In Figure~\ref{fig:different_threshold}, we examine \textsc{DePA}’s average F1-score across various values of $T$. The highest F1-score of 0.45 occurs at $T = 1.9$, although our default setting of $T=1.5$ delivers comparable results.

\begin{table*}
\caption{F1 Score of each detection methods.}
\label{fig:detect-posioned-dataset}
\centering
\begin{tabular}{clccccc}
\hline
\multirow{2}{*}{Dataset} &
  \multicolumn{1}{c}{\multirow{2}{*}{\begin{tabular}[c]{@{}c@{}}Poisoning \\ Method\end{tabular}}} &
  \multirow{2}{*}{\textsc{DePA}} &
  \multicolumn{2}{c}{ONION(CodeGPT)} &
  \multicolumn{2}{c}{ONION(CodeLlama)} \\ \cline{4-7} 
 &
  \multicolumn{1}{c}{} &
   &
  \begin{tabular}[c]{@{}c@{}}LLM \\ tokenizer\end{tabular} &
  \begin{tabular}[c]{@{}c@{}}Python \\ tokenizer\end{tabular} &
  \begin{tabular}[c]{@{}c@{}}LLM \\ tokenizer\end{tabular} &
  \begin{tabular}[c]{@{}c@{}}Python \\ tokenizer\end{tabular} \\ \hline
\multirow{9}{*}{MBPP}      & 1-Fixed   & \textbf{0.28} & 0.09 & 0.09 & 0.17 & 0.09 \\ \cline{2-7} 
                           & 1-Grammar & \textbf{0.27} & 0.09 & 0.09 & 0.18 & 0.09 \\ \cline{2-7} 
                           & 2-Fixed   & \textbf{0.29} & 0.09 & 0.09 & 0.07 & 0.09 \\ \cline{2-7} 
                           & 2-Grammar & \textbf{0.25} & 0.09 & 0.09 & 0.17 & 0.09 \\ \cline{2-7} 
                           & Random-1  & \textbf{0.28} & 0.09 & 0.09 & 0.16 & 0.09 \\ \cline{2-7} 
                           & Random-3  & \textbf{0.30} & 0.09 & 0.09 & 0.12 & 0.09 \\ \cline{2-7} 
                           & Random-5  & \textbf{0.29} & 0.09 & 0.09 & 0.08 & 0.09 \\ \cline{2-7} 
                           & Random-10 & \textbf{0.35} & 0.09 & 0.09 & 0.07 & 0.09 \\ \cline{2-7} 
                           & Random-20 & \textbf{0.41} & 0.09 & 0.09 & 0.06 & 0.09 \\ \hline
\multirow{4}{*}{HumanEval} & 1-Fixed   & \textbf{0.27} & 0.10 & 0.09 & 0.18 & 0.09 \\ \cline{2-7} 
                           & 1-Grammar & \textbf{0.27} & 0.10 & 0.09 & 0.22 & 0.09 \\ \cline{2-7} 
                           & 2-Fixed   & \textbf{0.23} & 0.10 & 0.09 & 0.18 & 0.09 \\ \cline{2-7} 
                           & 2-Grammar & \textbf{0.19} & 0.10 & 0.09 & 0.18 & 0.09 \\ \hline
\multicolumn{2}{c}{Average}            & \textbf{0.28} & 0.09 & 0.09 & 0.14 & 0.09 \\ \hline
\end{tabular}
\end{table*}

\begin{table*}
\caption{The average accuracy of locating dead code snippets across 4 attack types.}
\label{fig:locate-dead-code-acc}
\centering
\begin{tabular}{clccccc}
\hline
\multirow{2}{*}{Dataset} &
  \multicolumn{1}{c}{\multirow{2}{*}{\begin{tabular}[c]{@{}c@{}}Poisoning\\ Method\end{tabular}}} &
  \multirow{2}{*}{\textsc{DePA}} &
  \multicolumn{2}{c}{\begin{tabular}[c]{@{}c@{}}ONION\\ (CodeGPT)\end{tabular}} &
  \multicolumn{2}{c}{\begin{tabular}[c]{@{}c@{}}ONION\\ (CodeLlama)\end{tabular}} \\ \cline{4-7} 
 &
  \multicolumn{1}{c}{} &
   &
  \begin{tabular}[c]{@{}c@{}}LLM\\ tokenizer\end{tabular} &
  \begin{tabular}[c]{@{}c@{}}Python\\ tokenizer\end{tabular} &
  \begin{tabular}[c]{@{}c@{}}LLM\\ tokenizer\end{tabular} &
  \begin{tabular}[c]{@{}c@{}}Python\\ tokenizer\end{tabular} \\ \hline
\multirow{9}{*}{MBPP}          & 1-Fixed   & \textbf{0.98} & 0.17 & 0.39 & 0.07 & 0.26          \\ \cline{2-7} 
                               & 1-Grammer & \textbf{0.93} & 0.20 & 0.38 & 0.05 & 0.27          \\ \cline{2-7} 
                               & 2-Fixed   & \textbf{0.90} & 0.25 & 0.43 & 0.18 & 0.34          \\ \cline{2-7} 
                               & 2-Grammer & \textbf{0.96} & 0.26 & 0.42 & 0.20 & 0.32          \\ \cline{2-7} 
                               & Random-1  & \textbf{0.95} & 0.25 & 0.39 & 0.14 & 0.27          \\ \cline{2-7} 
                               & Random-3  & \textbf{0.71} & 0.52 & 0.57 & 0.40 & 0.57          \\ \cline{2-7} 
                               & Random-5  & \textbf{0.72} & 0.65 & 0.67 & 0.56 & 0.71          \\ \cline{2-7} 
                               & Random-10 & 0.76          & 0.78 & 0.79 & 0.75 & \textbf{0.83} \\ \cline{2-7} 
                               & Random-20 & 0.86          & 0.88 & 0.88 & 0.86 & \textbf{0.91} \\ \hline
\multirow{4}{*}{HumanEval}     & 1-Fixed   & \textbf{1.00} & 0.16 & 0.34 & 0.05 & 0.19          \\ \cline{2-7} 
                               & 1-Grammer & \textbf{1.00} & 0.24 & 0.30 & 0.09 & 0.19          \\ \cline{2-7} 
                               & 2-Fixed   & \textbf{0.92} & 0.24 & 0.39 & 0.13 & 0.26          \\ \cline{2-7} 
                               & 2-Grammer & \textbf{0.98} & 0.21 & 0.32 & 0.14 & 0.24          \\ \hline
\multirow{4}{*}{MathOA-Python} & 1-Fixed   & \textbf{0.92} & 0.13 & 0.32 & 0.04 & 0.13          \\ \cline{2-7} 
                               & 1-Grammer & \textbf{0.89} & 0.17 & 0.34 & 0.07 & 0.14          \\ \cline{2-7} 
                               & 2-Fixed   & \textbf{0.64} & 0.19 & 0.38 & 0.16 & 0.20          \\ \cline{2-7} 
                               & 2-Grammer & \textbf{0.82} & 0.21 & 0.35 & 0.16 & 0.22          \\ \hline
\multirow{4}{*}{APPS}          & 1-Fixed   & \textbf{0.74} & 0.09 & 0.21 & 0.02 & 0.11          \\ \cline{2-7} 
                               & 1-Grammer & \textbf{0.83} & 0.14 & 0.21 & 0.03 & 0.10          \\ \cline{2-7} 
                               & 2-Fixed   & \textbf{0.62} & 0.15 & 0.23 & 0.07 & 0.13          \\ \cline{2-7} 
                               & 2-Grammer & \textbf{0.79} & 0.15 & 0.22 & 0.08 & 0.13          \\ \hline
\multicolumn{2}{c}{Average}                & \textbf{0.85} & 0.29 & 0.41 & 0.20 & 0.31 \\ \hline
\end{tabular}
\end{table*}


\paragraph{Detection Speed}
Across all test datasets, \textsc{DePA} shows a clear advantage in detection speed. As reported in Table~\ref{fig:detect-performance}, \textsc{DePA} averages 88.16 samples per minute for four code dataset, demonstrating superior performance. In comparison, ONION(CodeGPT) processes 54.26 samples per minute, while ONION(CodeLlama) averages only 3.71. Table~\ref{fig:5_poisoned_dataset_processing_time} further confirms that \textsc{DePA} is the fastest in three out of four datasets, whereas ONION(CodeLlama) is the slowest, indicating ONION’s constraints in code-related tasks. These findings underscore \textsc{DePA}’s strengths not only in detection accuracy but also in processing speed.

\begin{table}[]
\caption{Detect performance of each detection methods.}
\label{fig:detect-performance}
\small	
\centering
\begin{tabular}{cccc}
\hline
Tasks/min &
  \textsc{DePA} &
  \begin{tabular}[c]{@{}c@{}}ONION\\ (CodeGPT)\end{tabular} &
  \begin{tabular}[c]{@{}c@{}}ONION\\ (CodeLlama)\end{tabular} \\ \hline
MBPP      & \textbf{149.46} & 120.49 & 9.10 \\ \hline
HumanEval & \textbf{129.47} & 46.35  & 2.37 \\ \hline
\begin{tabular}[c]{@{}c@{}}MathQA-\\ Python\end{tabular} &
  \textbf{68.23} &
  36.06 &
  2.92 \\ \hline
APPS      & \textbf{5.47}   & 14.13  & 0.43 \\ \hline
Average   & \textbf{88.16}  & 54.26  & 3.71 \\ \hline
\end{tabular}
\end{table}

\begin{table}[]
\caption{Detection of 5\% poisoned dataset processing time (unit: seconds).}
\label{fig:5_poisoned_dataset_processing_time}
\small
\centering
\begin{tabular}{cccc}
\hline
 &
  \textsc{DePA} &
  \begin{tabular}[c]{@{}c@{}}ONION\\ (CodeGPT)\end{tabular} &
  \begin{tabular}[c]{@{}c@{}}ONION\\ (CodeLlama)\end{tabular} \\ \hline
MBPP      & \textbf{22} & 24            & 316   \\ \hline
HumanEval & \textbf{10} & \textbf{10}   & 203   \\ \hline
\begin{tabular}[c]{@{}c@{}}MathQA-\\ Python\end{tabular} &
  \textbf{944} &
  1787 &
  22068 \\ \hline
APPS      & 4804        & \textbf{1860} & 61116 \\ \hline
\end{tabular}
\end{table} 

\begin{figure}
    \centering
    \includegraphics[width=1\linewidth]{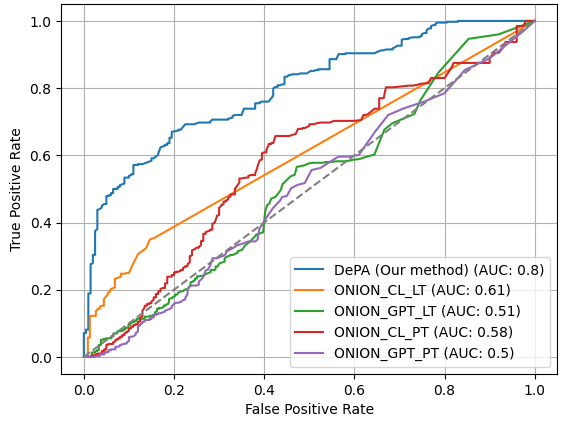}
    \caption{ROC curves of each detection methods (\textit{CL} refers to CodeLlama, \textit{GPT} indicates CodeGPT, \textit{LT} stands for the LLM Tokenizer, and \textit{PT} represents the Python Tokenizer.).}
    \label{fig:AUROC}
\end{figure}

\paragraph{AUROC}
Figure~\ref{fig:AUROC} shows the ROC curves for various detection methods. \textsc{DePA} notably outperforms the ONION baselines, reaching an AUROC of 0.80—indicating robust discriminative capability between poisoned (positive) and clean (negative) samples. By contrast, ONION(CodeGPT) achieves only 0.51 and 0.50 under both the CodeGPT and Python tokenizers, and ONION(CodeLlama) attains 0.61 and 0.58 in each tokenization setting.

\section{Discussion}
\paragraph{Adaptive Attack} An attacker may anticipate the use of \textsc{DePA}, leading us to examine an adaptive attack scenario. Since \textsc{DePA} relies on Equation~\ref{DePA(i)} for detection, one straightforward adversarial strategy is to craft dead code that slips past this threshold. Specifically, following \citet{wan2022you}, an attacker could use a genetic algorithm (GA)~\citep{man1996genetic} to generate complex grammar triggers designed to evade Equation~\ref{DePA(i)}. We applied such a poisoning attack to the MBPP dataset with a 5\% poisoning rate, using a population size of 100 and running for 20 iterations.

\begin{figure}
    \centering
    \includegraphics[width=1\linewidth]{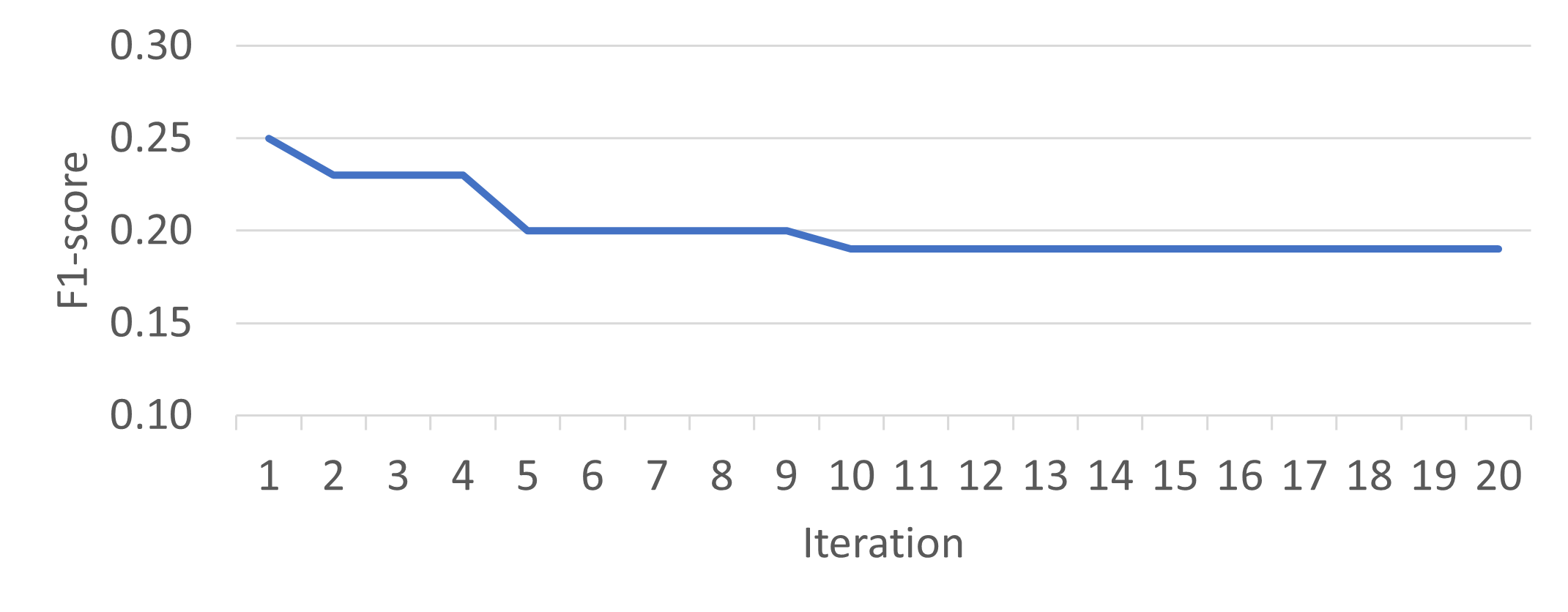}
    \caption{F1-scores of the varying number of iterations for GA in generating triggers that evade detection.}
    \label{fig:genetic_algo_experiment}
\end{figure}

As Figure~\ref{fig:genetic_algo_experiment} shows, the F1-score stabilized at 0.19 after 10 iterations. We then tested \textsc{DePA}, ONION(CodeGPT), and ONION(CodeLlama). Table~\ref{fig:genetic_algo_results} indicates that the detection accuracy of \textsc{DePA} fell to 0.19, while ONION(CodeGPT) and ONION(CodeLlama) dropped to 0.10 and 0.05, respectively. For dead code localization, \textsc{DePA} achieved 0.70, ONION(CodeGPT) 0.26, and ONION(CodeLlama) 0.22.

These findings suggest that although the genetic algorithm does not guarantee the absolute worst-case combination, it can efficiently discover near-optimal triggers that diminish the performance of both \textsc{DePA} and ONION-based methods. Nonetheless, detection remains viable, indicating that \textsc{DePA} maintains a degree of resilience against adaptive attacks.

\begin{table}[]
\caption{GA attack results of each detection methods.}
\label{fig:genetic_algo_results}
\small
\centering
\begin{tabular}{cccc}
\hline
 &
  \textsc{DePA} &
  \begin{tabular}[c]{@{}c@{}}ONION\\ (CodeGPT)\end{tabular} &
  \begin{tabular}[c]{@{}c@{}}ONION\\ (CodeLlama)\end{tabular} \\ \hline
\begin{tabular}[c]{@{}c@{}}Detection\\ F1-score\end{tabular} &
  \textbf{0.19} &
  0.10 &
  0.05 \\ \hline
\begin{tabular}[c]{@{}c@{}}Locating Dead \\ Code Accuracy\end{tabular} &
  \textbf{0.70} &
  0.26 &
  0.22 \\ \hline
\end{tabular}
\end{table}


    
\paragraph{Locating Poisoned Segment}
Regarding poisoned segment localization, \textsc{DePA} demonstrates 44-65\% improvement over baseline methods. Unlike ONION, which detects anomalies at the word level, \textsc{DePA} operates at the line level. As illustrated in Figure~\ref{fig:dead_code_localization_flow}, the second and third lines contain inserted dead code. \textcolor{red}{Red} text indicates correctly identified dead code, while \textcolor{blue}{blue} text marks false positives. By focusing on entire lines, \textsc{DePA} enhances localization accuracy.

However, this greater accuracy brings potential trade-offs. Since \textsc{DePA} uses line-level perplexity, it can still produce false positives against highly covert poisoning techniques—such as those modifying variables or function names \citep{sun2023backdooring, yang2024stealthy}. Future research should refine perplexity-based detection and incorporate additional features, including static analysis and syntax rule checks, to reduce false positives.

\begin{figure}
\small
\centering
\begin{tabular}{|l|l|}
\hline
\multicolumn{1}{|c|}{\textsc{DePA}} & \multicolumn{1}{c|}{ONION}          \\ \hline
def is\_not\_prime(n):     & def is\_{\color[HTML]{00B0F0}not}\_prime(n):              \\
\hspace{8pt} {\color[HTML]{FE0000} while random() \textgreater{}= 68:} & \hspace{8pt} while {\color[HTML]{FE0000}random}() {\color[HTML]{FE0000}\textgreater{}=} 68: \\
\hspace{16pt} return n                   & \hspace{16pt} {\color[HTML]{FE0000}return} n                            \\
\hspace{8pt} if n == 2 or n == 3:       & \hspace{8pt} if n == 2 or n == 3:                \\
\hspace{16pt} {\color[HTML]{00B0F0} return False}               & \hspace{16pt} return {\color[HTML]{00B0F0}False} \\
\hspace{8pt} for i in range(2, int(...  & \hspace{8pt} {\color[HTML]{00B0F0}for} i {\color[HTML]{00B0F0}in} range(2, {\color[HTML]{00B0F0}int}(...           \\
\hspace{16pt} ...                        & \hspace{16pt} ...                                 \\ \hline
\end{tabular}
\caption{Locating dead code via \textsc{DePA} and ONION.}
\label{fig:dead_code_localization_flow}
\end{figure}

\paragraph{Static Dead Code Detection Tools}
An alternative approach to detecting dead code is to use existing static analysis tools. For Python, tools like Vulture\footnote{Vulture (\url{https://github.com/jendrikseipp/vulture})} and Pylint\footnote{Pylint (\url{https://github.com/pylint-dev/pylint})} focus on locating unused variables, functions, and classes. However, they can only detect issues in a static context, whereas dead code can also emerge under conditions that never occur or loops that never run—situations that require runtime information to detect.

As shown in Figure~\ref{fig:comparison_depa_and_static_tools}, we consider a detection successful if Vulture or Pylint classifies the dead code as \textit{dead} or \textit{unreachable}. However, neither tool successfully flags the dead code described in \citet{ramakrishnan2022backdoors} and \citet{wan2022you}. In particular, the attack from \citet{ramakrishnan2022backdoors} uses \texttt{Exception}; Pylint noted that \texttt{Exception} was too generic but did not mark the snippet as dead or unreachable.

In contrast, \textsc{DePA} relies on a Code LLM rather than predefined coding rules. Similar to models trained on natural language, a Code LLM learns code properties through training. It can thus spot \textit{unreasonable} segments that would never execute at runtime—thereby overcoming the limitations of static analysis tools.

\begin{table}[]
\caption{Comparing \textsc{DePA} and static code analysis.}
\label{fig:comparison_depa_and_static_tools}
\small
\centering
\begin{tabular}{cccc}
\hline
          & \textsc{DePA} & Vulture & Pylint \\ \hline
1-Fixed   & \textbf{0.98}              & 0.00    & 0.00   \\ \hline
1-Grammer & \textbf{0.93}              & 0.00    & 0.00   \\ \hline
2-Fixed   & \textbf{0.90}              & 0.00    & 0.00   \\ \hline
2-Grammer & \textbf{0.96}              & 0.00    & 0.00   \\ \hline
\end{tabular}
\end{table}
    

\section{Conclusion}
In this paper, we introduced \textsc{DePA}, a novel method for detecting and cleansing dead code poisoning in code generation datasets. Unlike traditional token-level perplexity approaches, DEPA leverages the structural characteristics of code by performing line-level perplexity analysis, enabling it to identify anomalous lines with greater precision. Our findings highlight the importance of incorporating structural and contextual properties of code into detection mechanisms, paving the way for more secure and reliable code generation systems. 

\clearpage
\newpage

\section*{Limitations}
\textsc{DePA} primarily focus on dead code poisoning attacks in Python, but \textsc{DePA} may not be able to be seamlessly generalized to all programming languages. For example, C++ uses semicolons to separate statements, allowing multiple commands on a single line. This structure could lead \textsc{DePA} to misidentify poisoned code. Additionally, Python follows specific coding standards like PEP8, which sometimes splits lengthy statements across multiple lines. Although dead code is usually short, \textsc{DePA} may struggle with accurate detection, increasing false positives and reducing effectiveness if the original code spans multiple lines. Future work should explore adaptations for diverse languages and coding styles.

\bibliography{anthology,custom}
\bibliographystyle{acl_natbib}

\appendix

\section{Algorithm}
\label{sec:appendix}

\begin{algorithm2e}
    \SetCommentSty{mycommfont}
    \newcommand{\mycommfont}[1]{\footnotesize\textcolor{blue}{#1}}
    \DontPrintSemicolon
    \small
    \caption{\textsc{DePA}}\label{algo}
    \textbf{Input:} $D$: (Dataset), $M$: (CodeLlama), $T$: (Threshold) \\
    \textbf{Output:} $Pred$: (Prediction Result) \\
    \SetKwProg{Fn}{Function}{}{}
    \SetKwFunction{codeDetect}{codeDetect}
    \Fn{\codeDetect{$task$}:}{
        $text, code \gets task$ \\
        $code\_lines \gets$ Split $code$ into lines. \\
        $score \gets \{\}$ \\
        \For{$line$ in $code\_lines$}{ \tcp*{Initialize line score}
            $score[line] \gets \{"value": 0, "cnt": 0\}$ \\
        }
        \For{$idx=1$ to $len(code\_lines)$}{ \tcp*{Calculate combination perplexity}
            $code\_part \gets$ Merge $code\_lines$ except line $idx$ \\
            $PPL \gets M.perplexity(text, code\_part)$ \\
            \For{$line$ in $code\_part$}{
                $score[line]["value"] += PPL$ \\
                $score[line]["cnt"] += 1$ \\
            }
        }
        $score\_list \gets []$ \\
        \For{$s$ in $score$}{ \tcp*{Calculate line average perplexity}
            $line\_avg \gets s["value"]/s["cnt"]$ \\
            $score\_list.append(pow(line\_avg,2))$ \\
        }
        $avg \gets sum(score\_list)/len(score\_list)$ \\
        $std \gets np.std(score\_list)$ \\
        \For{$s$ in $score\_list$}{ \tcp*{Detect toxic code line}
            \If{$s - avg > T*std$}{
                \textbf{Return} $True$ \\
            }
        }
        \textbf{Return} $False$ \\
    }
    $Pred \gets []$ \\
    \For{$task$ in $D$}{
        $Pred.append(\codeDetect(task))$ \\
    }
    \textbf{Return} $Pred$ \\
\end{algorithm2e}

\end{document}